\documentclass{article}
\usepackage{spconf,amsmath,epsfig}
\usepackage{times}
\usepackage{graphicx}
\usepackage{amssymb}
\usepackage{array} 
\usepackage{subfigure} 
\usepackage{multirow} 
\usepackage[misc]{ifsym} 

\let\OLDthebibliography\thebibliography
\renewcommand\thebibliography[1]{
  \OLDthebibliography{#1}
  \setlength{\parskip}{0pt}
  \setlength{\itemsep}{0pt plus 0.3ex}
}

\begin{document}\sloppy
 
\def\x{{\mathbf x}}
\def\L{{\cal L}}

\title{UNITS: Unsupervised Intermediate Training Stage \\for Scene Text Detection}
%

%


\name{Youhui Guo\textsuperscript{1,2}, Yu Zhou\textsuperscript{1,2,*\thanks{*Corresponding author} 
        \thanks{Supported by the Beijing Municipal Science \& Technology Commission (Z191100007119002), the Key Research Program of Frontier Sciences, CAS, Grant NO ZDBS-LY-7024.} }, Xugong Qin\textsuperscript{1,2}, Enze Xie\textsuperscript{1}, Weiping Wang\textsuperscript{1}
      }

\address{\textsuperscript{1}Institute of Information Engineering, Chinese Academy of Sciences, Beijing, China  \\
        \textsuperscript{2}School of Cyber Security, University of Chinese Academy of Sciences, Beijing, China  \\
        {\{guoyouhui, zhouyu, qinxugong, wangweiping\}@iie.ac.cn}, xez18@mails.tsinghua.edu.cn
        }


\maketitle

\begin{abstract}
Recent scene text detection methods are almost based on deep learning and data-driven.
Synthetic data is commonly adopted for pre-training due to expensive annotation cost.
However, there are obvious domain discrepancies between synthetic data and real-world data. 
It may lead to sub-optimal performance to directly adopt the model initialized by synthetic data in the fine-tuning stage. In this paper, we propose a new training paradigm for scene text detection, which introduces an \textbf{UN}supervised \textbf{I}ntermediate \textbf{T}raining \textbf{S}tage (UNITS) that builds a buffer path to real-world data and can alleviate the gap between the pre-training stage and fine-tuning stage. Three training strategies are further explored to perceive information from real-world data in an unsupervised way. 
With UNITS, scene text detectors are improved without introducing any parameters and computations during inference. Extensive experimental results show consistent performance improvements on three public datasets. 
\end{abstract}
\section{Introduction}
\label{sec:intro}

Scene text detection is a fundamental and crucial task in computer vision because it is an important step in many practical applications, such as scene text recognition\cite{qz1,qz3}, image/video understanding, and text visual question answering \cite{zgy1}. 

\begin{figure}[!t]
\centering 
\includegraphics[width=0.48\textwidth]{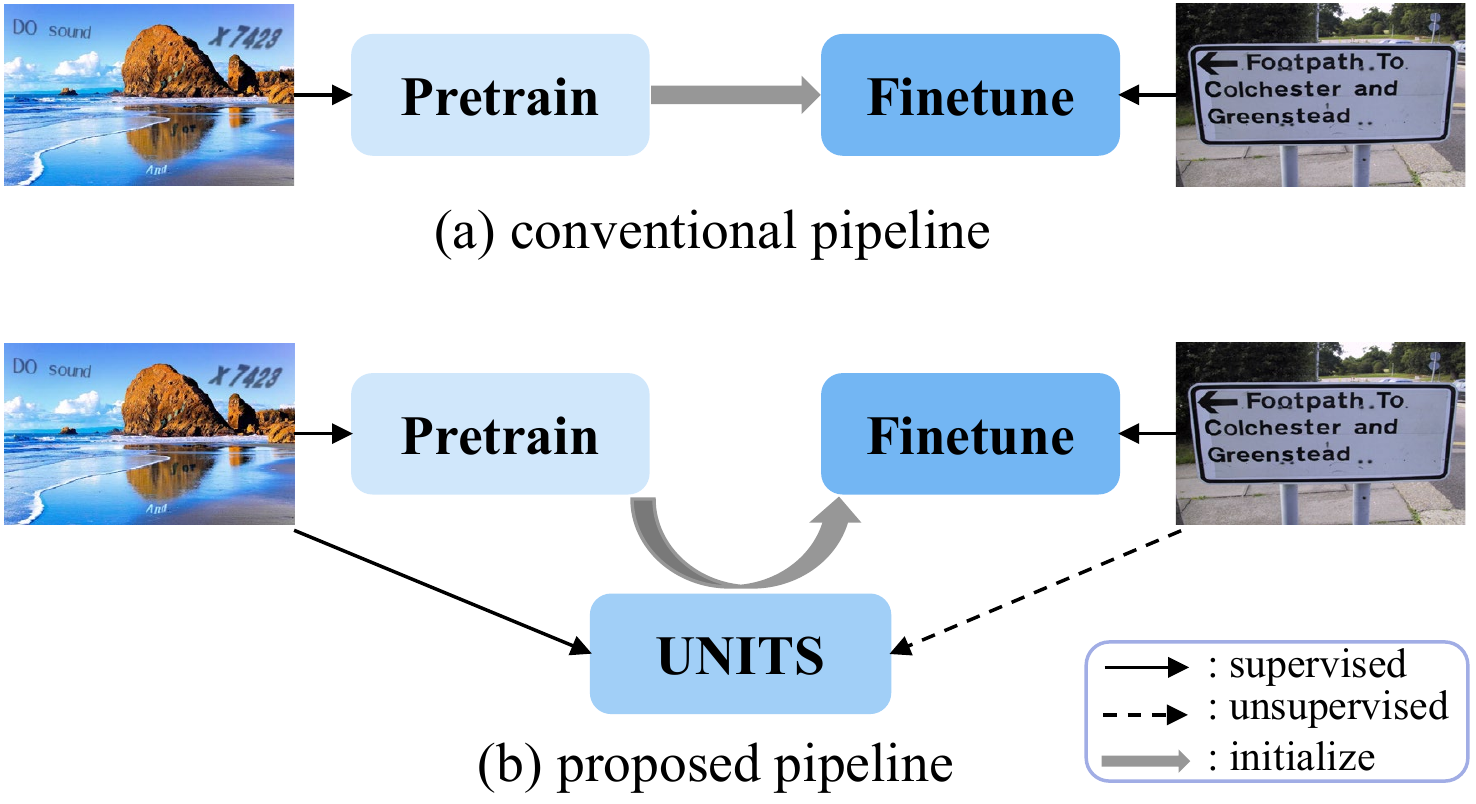}
\vspace{-0.7cm}
\caption{Comparison between the conventional training pipeline and our proposed pipeline for scene text detection. (a): The conventional pipeline consists of the pre-training stage and the fine-tuning stage. (b): We insert an unsupervised intermediate training stage (UNITS) to enable the pre-trained model to obtain information from real-world data.} 
\label{pipeline}
\vspace{-0.4cm}
\end{figure}

In recent years, deep learning based methods have been the mainstream in scene text detection, which require a large amount of data for training. However, labeling large sets of training images is extremely expensive and time-consuming. Most methods adopt synthetic data \cite{SynthText}, which is easy to obtain and the annotations can be freely and automatically generated, to pre-train the models before fine-tuning on real-world data for better performance as shown in Figure \ref{pipeline} (a). Text instances are born with large variations in shapes, colors, fonts, sizes, orientations, and blend naturally with the background, resulting in a large domain discrepancy between synthetic data and real-world data. 
It may be sub-optimal to initialize the model directly using the pre-trained parameters on synthetic data when fine-tuning on real-world data.

Most of the existing methods are dedicated to designing a better model, while few consider the data inconsistency between the pre-training stage and fine-tuning stage. 
Since 2017, the \textit{``pre-train and fine-tune''} paradigm dominates in natural language processing (NLP) tasks. Recently, a new paradigm dubbed \textit{``pre-train, prompt, and predict''} is proposed in which downstream tasks are reformulated to look more like those solved during pre-training with the help of a textual \textit{prompt}. 
Similarly, we propose a new training paradigm for scene text detection which includes an unsupervised intermediate training stage (UNITS) that can bridge the pre-training stage and fine-tuning stage and introduce information from real-world data into the pre-trained model. Inspired by the semi-supervised learning \cite{CPS,PseudoSeg,qxg2}, we design several unsupervised training strategies that only use the synthetic data and the unlabeled real-world data in UNITS. 
Specifically, for each unlabeled real image, we augment it to obtain an augmented version, and use the prediction of the original image to generate pseudo labels to supervise the prediction of the augmented image. In this way, the pre-trained model perceives the text information from the real-world data and provides a better initialization for the fine-tuning stage. Meanwhile, synthetic data is used to maintain the detection capability of the pre-trained model. Our proposed training pipeline with UNITS is illustrated in Figure \ref{pipeline} (b).

The proposed training paradigm is not constrained to one specific scene text detection model. We conduct extensive experiments on several classic scene text detection methods, e.g., EAST\cite{EAST}, PSENet\cite{PSENet}, PAN\cite{PAN} and DB\cite{DB} to verify the effectiveness and the generalization of the proposed method. Consistent performance improvements are obtained across different datasets and different scene text detectors.

The contributions of this work are summarized as follows:
\vspace{-0.6cm}
\begin{itemize}
\setlength{\itemsep}{-3pt}
\item A new training paradigm is proposed for scene text detection which includes an unsupervised intermediate training stage that can bridge the pre-training stage and fine-tuning stage, and no extra computations or model parameters are involved in inference. 

\item In the unsupervised intermediate training stage, we explore a class of unsupervised training strategies that only use synthetic data and unlabeled real-world data.

\item 
Experiments show that our method can appreciably improve the performance of several scene text detectors, e.g., DB \cite{DB} (83.8\%$\to$85.1\%), EAST \cite{EAST} (85.3\%$\to$86.5\%), PSENet \cite{PSENet} (81.2\%$\to$82.1\%), and PAN \cite{PAN} (82.5\%$\to$82.9\%) on ICDAR2015. Similar performance improvements can also be achieved on TotalText and MSRA-TD500 datasets. 
\end{itemize}

\section{Related Work}
With the revival of deep learning, the majority of recent scene text detectors are based on deep neural networks. These methods can be roughly classified into two categories: bottom-up methods and top-down methods.

\textbf{Bottom-up methods} detect the fundamental elements and group these elements into the final detection results. Some methods treat scene text detection as a semantic segmentation task and predict some auxiliary information to better differentiate pixels belonging to different text regions. PSENet \cite{PSENet} proposes a progressive scale expansion algorithm that can separate the dense text instances and detect text instances with arbitrary shapes. DB \cite{DB} performs the binarization process in a segmentation network with a differentiable binarization module to simplify post-processing.
Some methods \cite{cyd1,TextSnake,CRAFT,DRRG} first detect individual text parts or characters, and group them into texts with some special post-processing steps. 
CRAFT \cite{CRAFT} detects the text instances by exploring each character and affinity between characters. DRRG \cite{DRRG} constructs each text instance by a series of ordered rectangular components and utilizes graph convolution network to reason the relations of those components.

\textbf{Top-down methods} \cite{TextBoxes++,RRPN,EAST,gyh1} usually follow the general object detection methods and directly output the word/line-level detection results. 
TextBoxes++ \cite{TextBoxes++} utilizes quadrilateral regression and RRPN \cite{RRPN} proposes rotation region proposal to detect multi-oriented text. 
Differently, EAST \cite{EAST} is an anchor-free method that directly detects the quadrangles of words in a pixel-level manner. However, most of them show limited representation for irregular shapes, such as curved shapes.
Different from methods that focus on specific model design, we focus on the incoherence of the two stages of pre-training and fine-tuning caused by domain discrepancies between synthetic data and real-world data. The proposed method generalizes well to various text detection methods.


\textbf{Domain adaptation} aims to reduce the domain gap between training and testing data. There are also some methods \cite{zgy2,cyd2,GA-DAN,Synthetic-to-Real} to solve the domain adaptation problem in scene text detection. GA-DAN \cite{GA-DAN} converts a source-domain image into multiple images of different spatial views as in target domain. Wu et al. \cite{Synthetic-to-Real} aims at the serious domain difference between synthetic data and real-world data, and proposes a synthetic-to-real domain adaptation method for scene text detection, which transfers knowledge from synthetic data to real-world data.
In this work, we focus on how to use unlabeled real-world data to improve the pre-trained model to obtain better initialization and final performance during fine-tuning.

\section{Proposed Method}
\subsection{Overview}
\begin{figure}[!th]
\centering 
\includegraphics[width=0.36\textwidth]{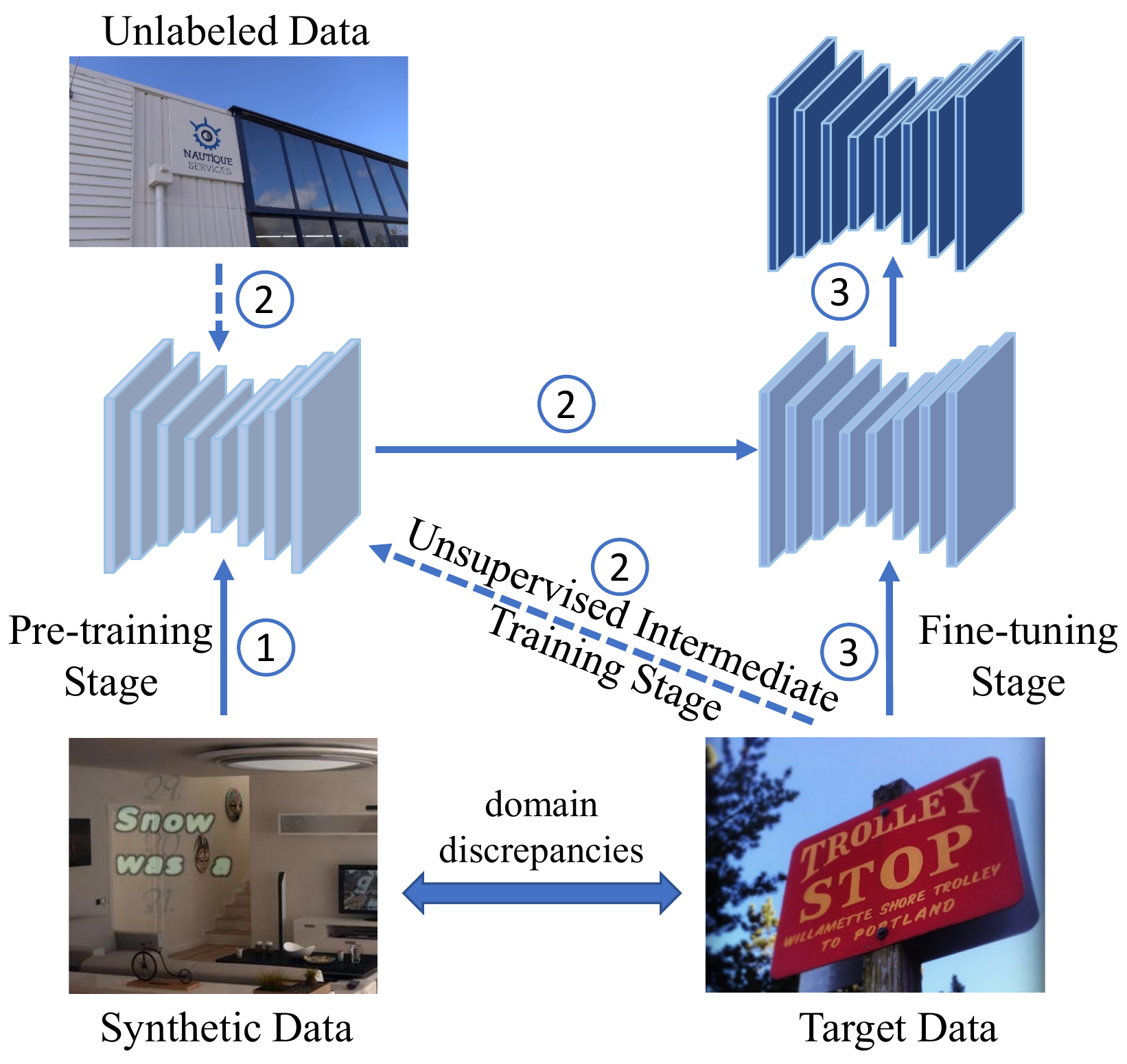}
\vspace{-0.5cm}
\caption{The illustration of our proposed training paradigm. The three training stages ``\textit{Pre-training, UNITS, and Fine-tuneing}'' are marked with serial numbers respectively.} 
\label{architecture}
\vspace{-0.5cm}
\end{figure}

In the training process of most previous scene text detection methods, the model is first trained using synthetic data (pre-training stage) and is trained on the target dataset (fine-tuning stage) with the initialization of the pre-trained model. In this section, we introduce a new training paradigm that includes an unsupervised intermediate training stage which bridges the pre-training stage and fine-tuning stage. The whole training paradigm is illustrated in Figure \ref{architecture}. Next, we first introduce the unsupervised intermediate training stage in detail. At the end of this section, we introduce how UNITS bridges the pre-training stage and fine-tuning stage.

\subsection{Unsupervised Intermediate Training Stage}

\begin{figure}[!th]
\centering 
\includegraphics[width=0.40\textwidth]{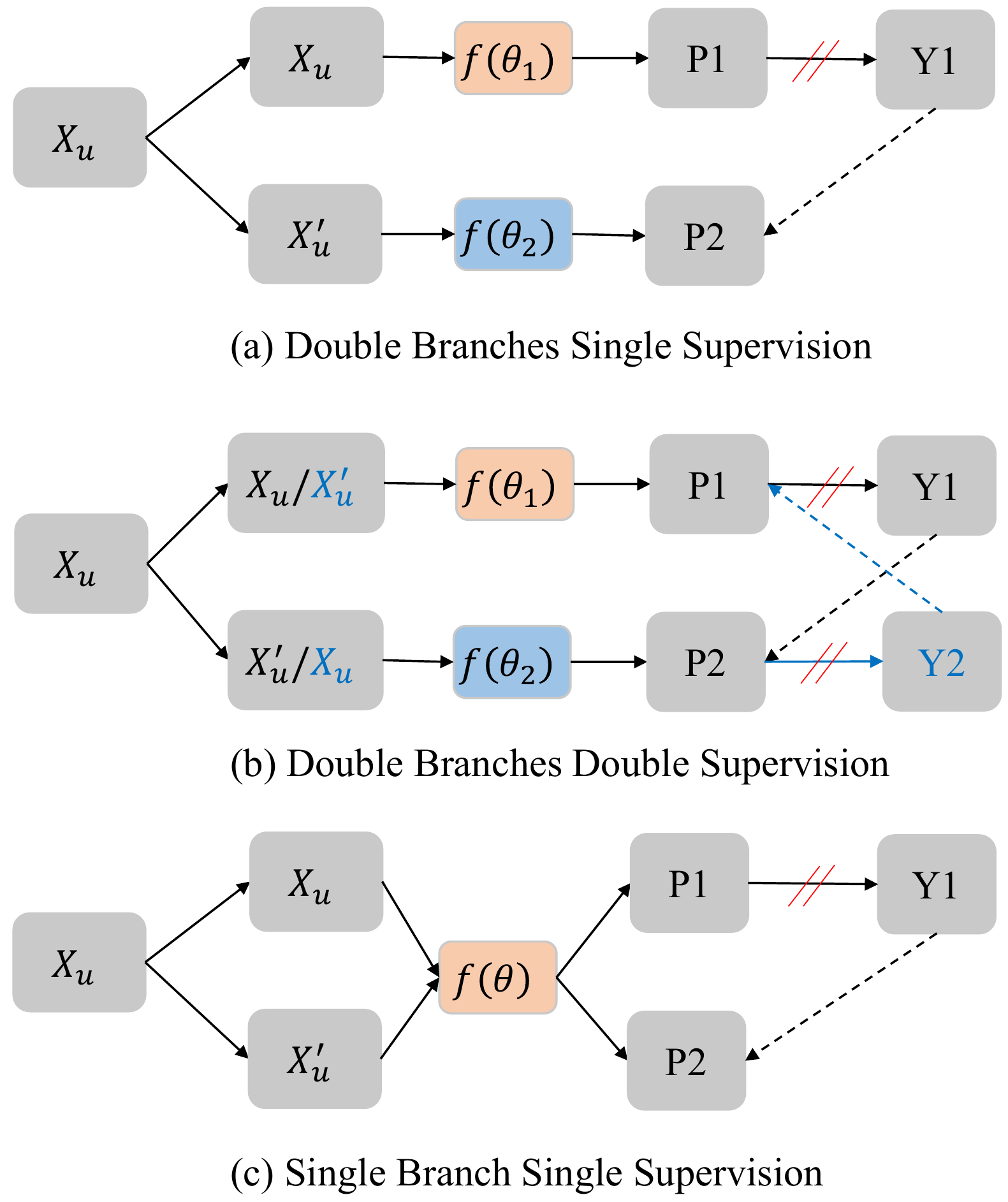}
\vspace{-0.5cm}
\caption{The illustration of our proposed training strategies for the unsupervised intermediate training stage. More details are introduced in the approach section. ``$\to$'' means forward operation and ``$\dashrightarrow$'' means loss supervision. ``//'' on ``$\to$'' means stop-gradient.} 
\label{study_1_arch}
\vspace{-0.5cm}
\end{figure}

Due to domain discrepancies between synthetic data and real-world data, directly applying a pre-trained model trained with synthetic data as initialization may not be optimal for the fine-tuning stage. To alleviate this, we explore three training strategies for the unsupervised intermediate training stage inspired by semi-supervised learning. With these training strategies, the pre-trained model can perceive text information from real-world data and provide better initialization compatible with the fine-tuning stage. The specific structures are shown in Figure~\ref{study_1_arch}, then we will introduce them separately.

\begin{table*}[!ht]
\centering
\small
\caption{Detection results with different unsupervised training strategies and comparisons with previous methods. ``P'', ``R'', and ``F'' indicate precision, recall, and f-measure respectively. ``*'' means our implementation.}
\label{tab:dts}
\begin{tabular}{|c|ccc|ccc|ccc|}
\hline
\multirow{2}{*}{Method}        & \multicolumn{3}{c|}{ICDAR2015}                                                          & \multicolumn{3}{c|}{MSRA-TD500}                                                         & \multicolumn{3}{c|}{TotalText}                                                                      \\ \cline{2-10} 
                               & \multicolumn{1}{c|}{P}             & \multicolumn{1}{c|}{R}             & F             & \multicolumn{1}{c|}{P}             & \multicolumn{1}{c|}{R}             & F             & \multicolumn{1}{c|}{P}             & \multicolumn{1}{c|}{R}             & F                         \\ \hline
SegLink \cite{SegLink}         & \multicolumn{1}{c|}{76.8}          & \multicolumn{1}{c|}{73.1}          & 75.0          & \multicolumn{1}{c|}{86.0}          & \multicolumn{1}{c|}{70.0}          & 77.0          & \multicolumn{1}{c|}{-}             & \multicolumn{1}{c|}{-}             & -                         \\ 
RRPN \cite{RRPN}               & \multicolumn{1}{c|}{82.0}          & \multicolumn{1}{c|}{73.0}          & 77.0          & \multicolumn{1}{c|}{82.0}          & \multicolumn{1}{c|}{68.0}          & 74.0          & \multicolumn{1}{c|}{-}             & \multicolumn{1}{c|}{-}             & -                         \\ 
TextBoxes++ \cite{TextBoxes++} & \multicolumn{1}{c|}{87.8}          & \multicolumn{1}{c|}{78.5}          & 82.9          & \multicolumn{1}{c|}{-}             & \multicolumn{1}{c|}{-}             & -             & \multicolumn{1}{c|}{-}             & \multicolumn{1}{c|}{-}             & -                         \\ 
TextSnake \cite{TextSnake}     & \multicolumn{1}{c|}{84.9}          & \multicolumn{1}{c|}{80.4}          & 82.6          & \multicolumn{1}{c|}{83.2}          & \multicolumn{1}{c|}{73.9}          & 78.3          & \multicolumn{1}{l|}{82.7}          & \multicolumn{1}{l|}{74.5}          & \multicolumn{1}{l|}{78.4} \\ 
CRAFT \cite{CRAFT}             & \multicolumn{1}{c|}{\textbf{89.9}} & \multicolumn{1}{c|}{\textbf{84.3}} & \textbf{86.9} & \multicolumn{1}{c|}{88.2}          & \multicolumn{1}{c|}{78.2}          & 82.9          & \multicolumn{1}{l|}{87.6}          & \multicolumn{1}{l|}{79.9}          & \multicolumn{1}{l|}{83.6} \\ 
PSENet + STKM \cite{STKM}      & \multicolumn{1}{c|}{87.8}          & \multicolumn{1}{c|}{84.1}          & 85.9          & \multicolumn{1}{c|}{-}             & \multicolumn{1}{c|}{-}             & -             & \multicolumn{1}{c|}{86.32}         & \multicolumn{1}{c|}{78.4}          & 82.0                      \\ \hline
DB* \cite{DB}                             & \multicolumn{1}{c|}{89.7}          & \multicolumn{1}{c|}{78.7}          & 83.8          & \multicolumn{1}{c|}{91.5}          & \multicolumn{1}{c|}{81.1}          & 86.0          & \multicolumn{1}{c|}{88.4}          & \multicolumn{1}{c|}{82.2}          & 85.2                      \\ 
DB* (1800 epochs)                           & \multicolumn{1}{c|}{89.8}          & \multicolumn{1}{c|}{79.0}          & 84.0          & \multicolumn{1}{c|}{89.2}          & \multicolumn{1}{c|}{80.6}          & 84.7          & \multicolumn{1}{c|}{88.2}          & \multicolumn{1}{c|}{81.2}          & 84.6                      \\ 
DB + SBSS                      & \multicolumn{1}{c|}{89.2}          & \multicolumn{1}{c|}{79.4}          & 84.0          & \multicolumn{1}{c|}{\textbf{91.9}} & \multicolumn{1}{c|}{82.3}          & 86.9          & \multicolumn{1}{c|}{88.7}          & \multicolumn{1}{c|}{83.1}          & \textbf{85.8}             \\ 
DB + DBDS                      & \multicolumn{1}{c|}{\textbf{89.9}} & \multicolumn{1}{c|}{79.4}          & 84.3          & \multicolumn{1}{c|}{91.5}          & \multicolumn{1}{c|}{84.7}          & 88.0          & \multicolumn{1}{c|}{\textbf{89.7}} & \multicolumn{1}{c|}{82.0}          & 85.7                      \\ 
DB + DBSS                      & \multicolumn{1}{c|}{89.6}          & \multicolumn{1}{c|}{\textbf{81.0}} & 85.1          & \multicolumn{1}{c|}{91.0}          & \multicolumn{1}{c|}{\textbf{85.2}} & \textbf{88.1} & \multicolumn{1}{c|}{88.6}          & \multicolumn{1}{c|}{\textbf{83.2}} & \textbf{85.8}             \\ \hline
\end{tabular}
\vspace{-0.4cm}
\end{table*}

\vspace{-0.4cm}
\subsubsection{Double Branches Single Supervision}

The Double Branches Single Supervision (DBSS) consists of two branches with the same structure but different parameters. One branch inputs the unlabeled real image $X_u$ to get the prediction result $P1$, and the other branch inputs the augmented version of $X^{'}_{ u}$ to get the prediction result $P2$, we use $P1$ to construct a new pseudo-label $Y1$ to supervise $P2$. For unlabeled real-world data, only the branch $f(\theta_{2})$ performs parameter update. In the fine-tuning stage, we use the branch $f(\theta_{2})$ to initialize the model and train specific epochs for different scene text detection detectors.

\vspace{-0.4cm}
\subsubsection{Double Branches Double Supervision}

The Double Branches Double Supervision (DBDS) is a modification of DBSS. We hope that the branch $f(\theta_{2})$ can obtain information from the real-world data in DBSS, then we consider whether we can make these two branches interactively promote each other to obtain a better $f(\theta_{1} )$ and $f(\theta_{2})$. On the basis of DBSS, we swap $X_u$ and $X^{'}_{u}$, and update the parameters of the branch $f(\theta_{1})$. Since the two branches are completely symmetrical for training, we use the branch $f(\theta_{1})$ or branch $f(\theta_{2})$ to initialize the model in the fine-tuning stage.

\vspace{-0.4cm}
\subsubsection{Single Branch Single Supervision}

The implementation of Single Branch Single Supervision (SBSS) refers to PseudoSeg \cite{PseudoSeg} and UDA \cite{UDA}. The input unlabeled real image $X_u$ is augmented to obtain the image $X^{'}_{u}$, and both enter the model $f(\theta)$ to get the prediction results, then we use the prediction result of $X_u$ to construct a pseudo-label to supervise the prediction result of $X^{'}_{u}$. If we equate $f(\theta_{1} )$ to $f(\theta_{2})$ in DBDS, we can get SBSS. Because there is only one branch, so we use the branch $f(\theta)$ to initialize the model in the fine-tuning stage.

\subsection{Training with UNITS}

Parameters pre-trained with synthetic data are used to initialize each of the above-mentioned branches. In order to maintain the detection capabilities of these branches, we keep each branch supervised with labeled synthetic images. In this way, the model can maintain a certain detection capability with synthetic data as well as perceive information from real-world data. Then the obtained weights are used to initialize the model in the fine-tuning stage on the target dataset. The whole training objective is formulated as:
\begin{equation}
\mathcal{L} = \mathcal{L}_{det} + \mathcal{L}_{units},
\end{equation}
where $\mathcal{L}_{det}$ is the detection loss in each text detector for labeled synthetic images, $\mathcal{L}_{units}$ is the loss supervision in UNITS for unlabeled real-world images.


\begin{table}[!t]
\centering
\small
\caption{Detection results with different detectors.}
\label{tab:diff_dets}
\begin{tabular}{|c|c|c|ccc|}
\hline
\multirow{2}{*}{Detectors} & \multirow{2}{*}{Backbone}  & \multirow{2}{*}{DBSS} & \multicolumn{3}{c|}{ICDAR2015}                                                          \\ \cline{4-6} 
                           &                            &                       & \multicolumn{1}{c|}{P}             & \multicolumn{1}{c|}{R}             & F             \\ \hline
\multirow{2}{*}{DB \cite{DB} }        & \multirow{2}{*}{ResNet-50} & $\times$              & \multicolumn{1}{c|}{\textbf{89.7}} & \multicolumn{1}{c|}{78.7}          & 83.8          \\ \cline{3-6} 
                           &                            & $\surd$               & \multicolumn{1}{c|}{89.6}          & \multicolumn{1}{c|}{\textbf{81.0}} & \textbf{85.1} \\ \hline
\multirow{2}{*}{EAST \cite{EAST} }      & \multirow{2}{*}{ResNet-50} & $\times$              & \multicolumn{1}{c|}{85.9}          & \multicolumn{1}{c|}{84.6}          & 85.3          \\ \cline{3-6} 
                           &                            & $\surd$               & \multicolumn{1}{c|}{\textbf{87.5}} & \multicolumn{1}{c|}{\textbf{85.4}} & \textbf{86.5} \\ \hline
\multirow{2}{*}{PSENet \cite{PSENet}}    & \multirow{2}{*}{ResNet-50} & $\times$              & \multicolumn{1}{c|}{\textbf{86.4}} & \multicolumn{1}{c|}{76.6}          & 81.2          \\ \cline{3-6} 
                           &                            & $\surd$               & \multicolumn{1}{c|}{84.8}          & \multicolumn{1}{c|}{\textbf{79.5}} & \textbf{82.1} \\ \hline
\multirow{2}{*}{PAN \cite{PAN}}       & \multirow{2}{*}{ResNet-18} & $\times$              & \multicolumn{1}{c|}{86.1}          & \multicolumn{1}{c|}{\textbf{79.2}} & 82.5          \\ \cline{3-6} 
                           &                            & $\surd$               & \multicolumn{1}{c|}{\textbf{87.2}} & \multicolumn{1}{c|}{79.0}          & \textbf{82.9} \\ \hline
\end{tabular}
\end{table}

\section{Experiments}
 
\subsection{Datasets}
The datasets used for the experiments in this paper are
briefly introduced below:

\noindent \textbf{SynthText} consists of 800k synthetic images. \\
\noindent \textbf{TotalText} is a dataset that includes horizontal, oriented, and curved text. It consists of 1,255 training images and 300 testing images.\\
\noindent \textbf{ICDAR2015} contains 1000 images for training and 500 images for testing. The images are captured by Google Glass and the text incidentally appears in the scene.  \\
\noindent \textbf{MSRA-TD500} is an oriented dataset that includes English and Chinese text instances with a large aspect ratio in natural scenes. It contains 300 training images and 200 testing images. Following the previous methods \cite{DB,qxg1,qxg3}, we include extra 400 images from HUST-TR400 for training.

\subsection{Implementation Details}

We experiment on multiple scene text detectors, including EAST$\footnote{https://github.com/SakuraRiven/EAST}$, DB$\footnote{https://github.com/MhLiao/DB}$, PSENet$\footnote{https://github.com/whai362/pan\_pp.pytorch\label{pan_pp}}$ and PAN$\textsuperscript{\ref{pan_pp}}$.
For simplicity, we use the pre-trained models provided by the open-source code of these methods. If the initial learning rate of the pre-training stage is $lr$, the initial learning rate of UNITS is set to $0.1*lr$. And if the training epoch of the fine-tuning stage is $epoch$, the training epoch of UNITS is set to $0.5*epoch$. In the fine-tuning stage, all training settings are the same as the original methods. We implement our method based on PyTorch framework and all the models run on GeForce RTX-2080Ti.

\begin{table}[!t]
\begin{center}
\small
\caption{Ablation study of the magnitude of data augmentation used in UNITS on ICDAR2015.}
\label{east_aug}
\resizebox{0.48\textwidth}{!}{
\begin{tabular}{|c|ccc|ccc|}
\hline
\multirow{2}{*}{Augmentation Type} & \multicolumn{3}{c|}{EAST}                                    & \multicolumn{3}{c|}{DB}                                      \\ \cline{2-7} 
                                   & \multicolumn{1}{c|}{P}    & \multicolumn{1}{c|}{R}    & F    & \multicolumn{1}{c|}{P}    & \multicolumn{1}{c|}{R}    & F    \\ \hline
Baseline                           & \multicolumn{1}{c|}{85.9} & \multicolumn{1}{c|}{84.6} & 85.3 & \multicolumn{1}{c|}{\textbf{89.7}} & \multicolumn{1}{c|}{78.7} & 83.8 \\ \hline
Color Jitter                       & \multicolumn{1}{c|}{85.8} & \multicolumn{1}{c|}{84.7} & 85.3 & \multicolumn{1}{c|}{89.5}     & \multicolumn{1}{c|}{79.1} & 83.9    \\ \hline
Randon Rotate                      & \multicolumn{1}{c|}{\textbf{87.5}} & \multicolumn{1}{c|}{\textbf{85.4}} & \textbf{86.5} & \multicolumn{1}{c|}{89.6} & \multicolumn{1}{c|}{\textbf{81.0}} & \textbf{85.1} \\ \hline
\end{tabular}}
\end{center}
\vspace{-0.6cm}
\end{table}

\begin{table*}[!ht]
\centering
\small
\caption{The effectiveness of multiple data augmentations. ``Single Augmentation'' means random rotation, and ``Multiple Augmentations'' means random rotation, random crop, and random scale.}
\label{tab:mutil_aug}
\begin{tabular}{|c|ccc|ccc|ccc|}
\hline
\multirow{2}{*}{Method} & \multicolumn{3}{c|}{TotalText}                                                          & \multicolumn{3}{c|}{ICDAR2015}                                                          & \multicolumn{3}{c|}{MSRA-TD500}                                                         \\ \cline{2-10} 
                        & \multicolumn{1}{c|}{P}             & \multicolumn{1}{c|}{R}             & F             & \multicolumn{1}{c|}{P}             & \multicolumn{1}{c|}{R}             & F             & \multicolumn{1}{c|}{P}             & \multicolumn{1}{c|}{R}             & F             \\ \hline
Baseline                & \multicolumn{1}{c|}{88.4}          & \multicolumn{1}{c|}{82.2}          & 85.2          & \multicolumn{1}{c|}{89.7}          & \multicolumn{1}{c|}{78.7}          & 83.8          & \multicolumn{1}{c|}{91.5}          & \multicolumn{1}{c|}{81.1}          & 86.0          \\ \hline
Single Augmentation     & \multicolumn{1}{c|}{88.6}          & \multicolumn{1}{c|}{\textbf{83.2}} & \textbf{85.8} & \multicolumn{1}{c|}{\textbf{89.6}} & \multicolumn{1}{c|}{\textbf{81.0}} & \textbf{85.1} & \multicolumn{1}{c|}{91.0}          & \multicolumn{1}{c|}{\textbf{85.2}} & \textbf{88.1} \\ \hline
Multiple Augmentations  & \multicolumn{1}{c|}{\textbf{88.8}} & \multicolumn{1}{c|}{83.0}          & \textbf{85.8} & \multicolumn{1}{c|}{89.3}          & \multicolumn{1}{c|}{80.7}          & 84.8          & \multicolumn{1}{c|}{\textbf{92.4}} & \multicolumn{1}{c|}{83.0}          & 87.4          \\ \hline
\end{tabular}
\vspace{-0.4cm}
\end{table*}

\begin{table*}[!ht]
\centering
\small
\caption{Ablation study of single dataset v.s. multiple datasets in UNITS.}
\label{tab:mutil_data}
\begin{tabular}{|c|ccc|ccc|ccc|}
\hline
\multirow{2}{*}{Method} & \multicolumn{3}{c|}{TotalText}                                                          & \multicolumn{3}{c|}{ICDAR2015}                                                          & \multicolumn{3}{c|}{MSRA-TD500}                                                                     \\ \cline{2-10} 
                        & \multicolumn{1}{c|}{P}             & \multicolumn{1}{c|}{R}             & F             & \multicolumn{1}{c|}{P}             & \multicolumn{1}{c|}{R}             & F             & \multicolumn{1}{c|}{P}             & \multicolumn{1}{c|}{R}             & F                         \\ \hline
Baseline                & \multicolumn{1}{c|}{88.4}          & \multicolumn{1}{c|}{82.2}          & 85.2          & \multicolumn{1}{c|}{89.7}          & \multicolumn{1}{c|}{78.7}          & 83.8          & \multicolumn{1}{c|}{91.5}          & \multicolumn{1}{c|}{81.1}          & 86.0                      \\ \hline
Single Dataset          & \multicolumn{1}{c|}{88.6}          & \multicolumn{1}{c|}{\textbf{83.2}} & 85.8          & \multicolumn{1}{c|}{89.6}          & \multicolumn{1}{c|}{\textbf{81.0}} & \textbf{85.1} & \multicolumn{1}{c|}{91.0}          & \multicolumn{1}{c|}{\textbf{85.2}} & \textbf{88.1}             \\ \hline
Multiple Datasets       & \multicolumn{1}{c|}{\textbf{90.9}} & \multicolumn{1}{c|}{81.6}          & \textbf{86.0} & \multicolumn{1}{c|}{\textbf{90.8}} & \multicolumn{1}{c|}{79.5}          & 84.8          & \multicolumn{1}{l|}{\textbf{91.6}} & \multicolumn{1}{l|}{82.1}          & \multicolumn{1}{l|}{86.6} \\ \hline
\end{tabular}
\vspace{-0.4cm}
\end{table*}

\subsection{Ablation Study}

\subsubsection{Different Training Strategies}
To verify the effectiveness of the three different training strategies in UNITS, we conduct experimental verification on DB. Following the open-source project, we conduct experiments on ICDAR2015, TotalText and MSRA-TD500. Specifically, we train 600 epochs in UNITS, then use the trained model as the initialization, and finally fine-tune 1200 epochs on the real-world dataset. The experimental results are shown in Table \ref{tab:dts}. It is worth mentioning that the baseline uses branch $f(\theta_{1})$ in DBSS to initialize the model in order to eliminate the effect of using synthetic data in UNITS. It can be seen that the three training strategies can bring performance improvements on three datasets, which verifies the effectiveness of our new training pipeline. The DBSS achieves the best performance among the three training strategies, so we adopt it as the default training strategy for the subsequent experiments.

Also, we fine-tune 1800 epochs for the baseline DB and the performances have not increased or even decreased due to overfitting, which validates that the performance improvements do not come from additional training epochs.

\begin{figure*}[t]
\centering 
\includegraphics[width=0.9\textwidth]{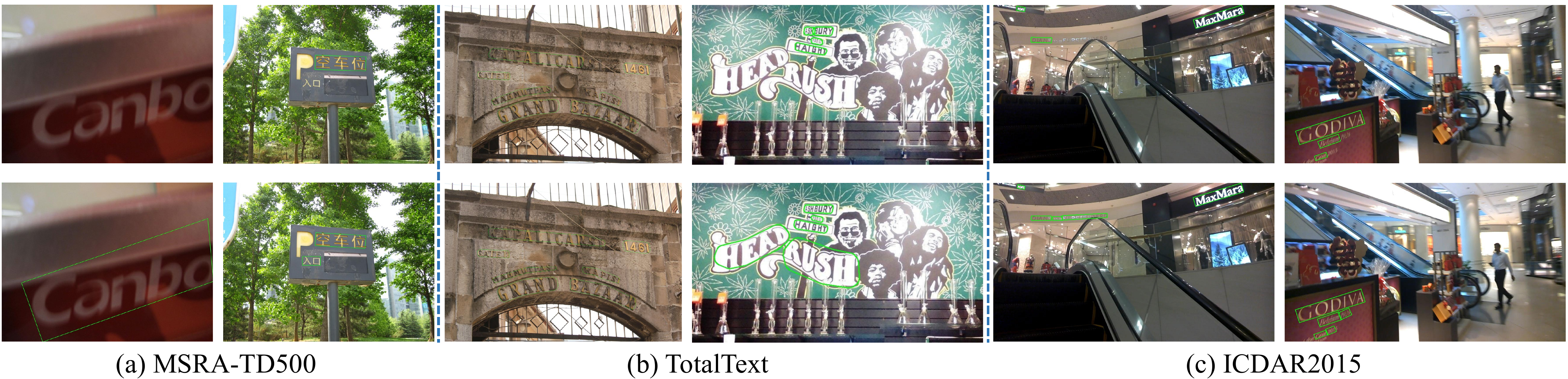}
\vspace{-0.4cm}
\caption{Qualitative results on different datasets. The upper and lower rows are the results of baseline and UNITS respectively.} 
\label{vis_last}
\vspace{-0.2cm}
\end{figure*}

\vspace{-0.2cm}
\subsubsection{Data Augmentation}
First, we verify whether weak augmentation or strong augmentation should be used in UNITS. Based on DB and EAST, we conduct experiments on ICDAR2015, and the training strategy is DBSS. The experimental results are shown in Table \ref{east_aug}. It can be seen that the performance is basically unchanged when weak augmentation (color jitter) is adopted. Because the input images of the two branches are basically the same and the output is almost the same, the provided supervision is roughly equivalent to supervision from pre-training with synthetic data alone. When strong augmentation (random rotation) is used, it can bring performance improvement of 1.2\% and 1.4\% on EAST and DB respectively. 

Then we explore the role of multiple data augmentations. On the basis of random rotation, we add two additional data augmentations: random crop and random scale. We conduct experiments on DB, and the experimental results are shown in Table \ref{tab:mutil_aug}. Both can bring close performance improvements on MSRA-TD500. The performances of single augmentation are slightly better than multiple augmentations on ICDAR2015 and TotalText. We use random rotation as the default augmentation setting.

\vspace{-0.3cm}
\subsubsection{Domain Discrepancies}
To verify whether our method can alleviate the domain discrepancies, we directly test the models of pre-training and UNITS on the test dataset. The model of UNITS is consistently superior to the model of pre-training, e.g., MSRA-TD500 (39.9\% $\to$ 44.6\%), ICDAR2015 (49.5\% $\to$ 52.6\%), and TotalText (49.8\% $\to$ 51.5\%) with DBSS.

The domain adaptation methods are to achieve the highest possible performance in the target domain. We use the labeled real-world data in DBSS to approximate the best domain adaptation result which can achieve 83.9\% in F-measure on MSRA-TD500. After fine-tuning, the F-measure increases to 86.5\% which is 0.5\% higher than baseline, and the slight performance improvement may come from strong data augmentation. However, the F-measure of DBSS increases from 44.6\% to 88.1\% after fine-tuning. It shows that better performance before fine-tuning does not mean better performance after fine-tuning. The domain adaptation methods migrate the model to the target domain as much as possible but lose the knowledge learned by pre-training on large-scale synthetic data. UNITS is a trade-off between them and serves as a good bridge between the pre-training and fine-tuning stages.

\vspace{-0.3cm}
\subsubsection{Multiple Datasets for UNITS}

As shown in Figure \ref{architecture}, we explore the possibility of using other unlabeled data. We use the existing public training datasets (ART, MLT2017, MSRA-TD500, HUST-TR400, and ICDAR2015) to construct a large dataset for UNITS training, denoted as UnlabeledDataset. By training UNITS on UnlabeledDataset, we can get an initialization model that is directly used to fine-tune all target datasets, which can reduce the time required to retrain for each dataset. Also, we use DBSS in UNITS and experiment on DB. The results are shown in Table \ref{tab:mutil_data}. More data can bring similar performance improvement for ICDAR2015 and TotalText, but the improvement is tiny for MSRA-TD500. Since MSRA-TD500 consists of mostly text instances with a large aspect ratio which occupy a small proportion in UnlabeledDataset.

\vspace{-0.4cm}
\subsection{Comparison on Different Scene Text Detectors}
To further verify the effectiveness of our proposed method, we conduct experiments on different scene text detectors. Specifically, we use DBSS as the training strategy in UNITS and the experimental dataset is ICDAR2015. The experimental results are shown in the Table \ref{tab:diff_dets}. With UNITS, UNITS achieve substantial improvements of 1.3\%, 1.2\%, 0.9\%, and 0.5\% in F-measure on DB, EAST, PSENet, and PAN respectively, which demonstrates that our method generalizes well across different detectors.

\vspace{-0.4cm}
\subsection{Comparisons with State-of-the-Art Methods}
We also compare DB with UNITS with state-of-the-art methods and the results are shown in Table \ref{tab:dts}. Our method can achieve better performance among the previous methods on MSRA-TD500 and TotalText. Especially, In particular, our method achieves 88.1\% in F-measure and is significantly higher than other methods on MSRA-TD500. For ICDAR2015, our method is lower than CRAFT \cite{CRAFT} which adopts additional character-level supervision. Some qualitative results on DB are displayed in Figure \ref{vis_last}. Compared with the baseline, our method can detect more easily missed texts and suppress some false positives similar to the text instances.

\section{Conclusion}
In this paper, we propose a simple and effective training paradigm for scene text detection in which an unsupervised intermediate training stage (UNITS) can bridge the pre-training stage and fine-tuning stage and introduce information from real-world data into the pre-trained model. The experiments conducted on various benchmarks verify the effectiveness of UNITS across various scene text detectors. 
In the future, we hope to explore more training strategies like unsupervised representation learning methods \cite{ldz1,ldz2,lxn2,luo2022exploring} or incremental learning methods \cite{yang2022rd} in UNITS and apply our training paradigm to more types of text detection methods.

\footnotesize
\bibliographystyle{IEEEbib}
\bibliography{icme2021template}

\begin{thebibliography}{10}

\bibitem{qz1}
Zhi Qiao, Yu~Zhou, and et~al.,
\newblock ``{SEED:} semantics enhanced encoder-decoder framework for scene text
  recognition,''
\newblock in {\em CVPR}, 2020, pp. 13525--13534.

\bibitem{qz3}
Zhi Qiao, Yu~Zhou, and et~al.,
\newblock ``Pimnet: {A} parallel, iterative and mimicking network for scene
  text recognition,''
\newblock in {\em {ACM} {MM}}, 2021, pp. 2046--2055.

\bibitem{zgy1}
Gangyan Zeng, Yuan Zhang, Yu~Zhou, and Xiaomeng Yang,
\newblock ``Beyond {OCR} + {VQA:} involving {OCR} into the flow for robust and
  accurate textvqa,''
\newblock in {\em {ACM} {MM}}, 2021, pp. 376--385.

\bibitem{SynthText}
Ankush Gupta, Andrea Vedaldi, and Andrew Zisserman,
\newblock ``Synthetic data for text localisation in natural images,''
\newblock in {\em CVPR}. 2016, {IEEE}.

\bibitem{CPS}
Xiaokang Chen, Yuhui Yuan, Gang Zeng, and Jingdong Wang,
\newblock ``Semi-supervised semantic segmentation with cross pseudo
  supervision,''
\newblock in {\em {CVPR}}, 2021, pp. 2613--2622.

\bibitem{PseudoSeg}
Yuliang Zou, Zizhao Zhang, Han Zhang, Chun{-}Liang Li, Xiao Bian, Jia{-}Bin
  Huang, and Tomas Pfister,
\newblock ``Pseudoseg: Designing pseudo labels for semantic segmentation,''
\newblock in {\em {ICLR}}, 2021.

\bibitem{qxg2}
Xugong Qin, Yu~Zhou, Dongbao Yang, and Weiping Wang,
\newblock ``Curved text detection in natural scene images with semi- and
  weakly-supervised learning,''
\newblock in {\em {ICDAR}}, 2019, pp. 559--564.

\bibitem{EAST}
Xinyu Zhou, Cong Yao, He~Wen, Yuzhi Wang, Shuchang Zhou, Weiran He, and Jiajun
  Liang,
\newblock ``{EAST:} an efficient and accurate scene text detector,''
\newblock in {\em {CVPR}}, 2017, pp. 2642--2651.

\bibitem{PSENet}
Wenhai Wang, Enze Xie, Xiang Li, Wenbo Hou, Tong Lu, Gang Yu, and Shuai Shao,
\newblock ``Shape robust text detection with progressive scale expansion
  network,''
\newblock in {\em {CVPR}}, 2019, pp. 9336--9345.

\bibitem{PAN}
Wenhai Wang, Enze Xie, and et~al.,
\newblock ``Efficient and accurate arbitrary-shaped text detection with pixel
  aggregation network,''
\newblock in {\em {ICCV}}, 2019, pp. 8439--8448.

\bibitem{DB}
Minghui Liao, Zhaoyi Wan, Cong Yao, Kai Chen, and Xiang Bai,
\newblock ``Real-time scene text detection with differentiable binarization,''
\newblock in {\em {AAAI}}, 2020, pp. 11474--11481.

\bibitem{cyd1}
Yudi Chen, Yu~Zhou, and et~al.,
\newblock ``Constrained relation network for character detection in scene
  images,''
\newblock in {\em {PRICAI}}, 2019, pp. 137--149.

\bibitem{TextSnake}
Shangbang Long, Jiaqiang Ruan, and et~al.,
\newblock ``Textsnake: {A} flexible representation for detecting text of
  arbitrary shapes,''
\newblock in {\em {ECCV}}, 2018, pp. 19--35.

\bibitem{CRAFT}
Youngmin Baek, Bado Lee, and et~al.,
\newblock ``Character region awareness for text detection,''
\newblock in {\em {CVPR}}, 2019, pp. 9365--9374.

\bibitem{DRRG}
Shi{-}Xue Zhang, Xiaobin Zhu, and et~al.,
\newblock ``Deep relational reasoning graph network for arbitrary shape text
  detection,''
\newblock in {\em {CVPR}}, 2020, pp. 9696--9705.

\bibitem{TextBoxes++}
Minghui Liao, Baoguang Shi, and Xiang Bai,
\newblock ``Textboxes++: {A} single-shot oriented scene text detector,''
\newblock {\em {IEEE} Trans. Image Process.}, vol. 27, no. 8, pp. 3676--3690,
  2018.

\bibitem{RRPN}
Jianqi Ma, Weiyuan Shao, and et~al.,
\newblock ``Arbitrary-oriented scene text detection via rotation proposals,''
\newblock {\em TMM}, vol. 20, no. 11, pp. 3111--3122, 2018.

\bibitem{gyh1}
Youhui Guo, Yu~Zhou, Xugong Qin, and Weiping Wang,
\newblock ``Which and where to focus: A simple yet accurate framework for
  arbitrary-shaped nearby text detection in scene images,''
\newblock in {\em {ICANN}}, 2021.

\bibitem{zgy2}
Gangyan Zeng, Yuan Zhang, and et~al.,
\newblock ``A cost-efficient framework for scene text detection in the wild,''
\newblock in {\em {PRICAI}}, 2021.

\bibitem{cyd2}
Yudi Chen, Wei Wang, and et~al.,
\newblock ``Self-training for domain adaptive scene text detection,''
\newblock in {\em {ICPR}}, 2020, pp. 850--857.

\bibitem{GA-DAN}
Fangneng Zhan, Chuhui Xue, and Shijian Lu,
\newblock ``{GA-DAN:} geometry-aware domain adaptation network for scene text
  detection and recognition,''
\newblock in {\em {ICCV}}, 2019, pp. 9104--9114.

\bibitem{Synthetic-to-Real}
Weijia Wu, Ning Lu, Enze Xie, Yuxing Wang, Wenwen Yu, Cheng Yang, and Hong
  Zhou,
\newblock ``Synthetic-to-real unsupervised domain adaptation for scene text
  detection in the wild,''
\newblock in {\em {ACCV}}, 2020, pp. 289--303.

\bibitem{SegLink}
Baoguang Shi, Xiang Bai, and et~al.,
\newblock ``Detecting oriented text in natural images by linking segments,''
\newblock in {\em {CVPR}}, 2017, pp. 3482--3490.

\bibitem{STKM}
Qi~Wan, Haoqin Ji, and Linlin Shen,
\newblock ``Self-attention based text knowledge mining for text detection,''
\newblock in {\em {CVPR}}, 2021, pp. 5983--5992.

\bibitem{UDA}
Qizhe Xie, Zihang Dai, and et~al.,
\newblock ``Unsupervised data augmentation for consistency training,''
\newblock in {\em NeurIPS}, 2020.

\bibitem{qxg1}
Xugong Qin, Yu~Zhou, Youhui Guo, Dayan Wu, and Weiping Wang,
\newblock ``Fc$^{2}$rn: {A} fully convolutional corner refinement network for
  accurate multi-oriented scene text detection,''
\newblock in {\em {ICASSP}}, 2021, pp. 4350--4354.

\bibitem{qxg3}
Xugong Qin, Yu~Zhou, and et~al.,
\newblock ``Mask is all you need: Rethinking mask {R-CNN} for dense and
  arbitrary-shaped scene text detection,''
\newblock in {\em {ACM} {MM}}, 2021, pp. 414--423.

\bibitem{ldz1}
Dezhao Luo, Chang Liu, and et~al.,
\newblock ``Video cloze procedure for self-supervised spatio-temporal
  learning,''
\newblock in {\em {AAAI}}, 2020, pp. 11701--11708.

\bibitem{ldz2}
Yuan Yao, Chang Liu, Dezhao Luo, Yu~Zhou, and Qixiang Ye,
\newblock ``Video playback rate perception for self-supervised spatio-temporal
  representation learning,''
\newblock in {\em {CVPR}}, 2020, pp. 6547--6556.

\bibitem{lxn2}
Xiaoni Li, Yu~Zhou, and et~al.,
\newblock ``Dense semantic contrast for self-supervised visual representation
  learning,''
\newblock in {\em {ACM} {MM}}, 2021, pp. 1368--1376.

\bibitem{luo2022exploring}
Dezhao Luo, Yu~Zhou, and et~al.,
\newblock ``Exploring relations in untrimmed videos for self-supervised
  learning,''
\newblock {\em ACM TOMM}, vol. 18, no. 1s, pp. 1--21, 2022.

\bibitem{yang2022rd}
Dongbao Yang, Yu~Zhou, and et~al.,
\newblock ``Rd-iod: Two-level residual-distillation-based triple-network for
  incremental object detection,''
\newblock {\em ACM TOMM}, vol. 18, no. 1, pp. 1--23, 2022.

\end{thebibliography}

\end{document}